%% file: ijcnn18-carestruct-main.tex
\documentclass[english,conference]{IEEEtran}
\usepackage[latin9]{inputenc}
\usepackage{array}
\usepackage{multirow}
\usepackage{amsmath}
\usepackage{amssymb}
\usepackage{graphicx}

\makeatletter

\providecommand{\tabularnewline}{\\}
\newcommand{\lyxdot}{.}

\usepackage{microtype}
\usepackage{flushend}

\makeatother

\usepackage{babel}
\begin{document}
\global\long\def\model{\mathsf{\mathtt{Resset}}}

\title{Resset: A Recurrent Model for Sequence of Sets with Applications to Electronic Medical Records}

\author{Phuoc Nguyen, Truyen Tran, Svetha Venkatesh\\
Applied AI Institute, Deakin University\\
\emph{\{phuoc.nguyen,truyen.tran,svetha.venkatesh\}@deakin.edu.au}}
\maketitle
\begin{abstract}
\input{abs.tex}
\end{abstract}

\global\long\def\xb{\boldsymbol{x}}
\global\long\def\yb{\boldsymbol{y}}
\global\long\def\hb{\boldsymbol{h}}
\global\long\def\Xcal{\mathcal{X}}
\global\long\def\Ucal{\mathcal{U}}
\global\long\def\Vcal{\mathcal{V}}
\global\long\def\Real{\mathbb{R}}
\global\long\def\mat{\text{mat}}
\global\long\def\thetab{\boldsymbol{\theta}}
\global\long\def\Wb{\boldsymbol{W}}
\global\long\def\bb{\boldsymbol{b}}
\global\long\def\cb{\boldsymbol{c}}
\global\long\def\fb{\boldsymbol{f}}
\global\long\def\gb{\boldsymbol{g}}
\global\long\def\ib{\boldsymbol{i}}
\global\long\def\ob{\boldsymbol{o}}
\global\long\def\ub{\boldsymbol{u}}
\global\long\def\vb{\boldsymbol{v}}
\global\long\def\db{\boldsymbol{d}}
\global\long\def\pb{\boldsymbol{p}}
\global\long\def\eb{\boldsymbol{e}}
\global\long\def\Db{\boldsymbol{D}}
\global\long\def\Ib{\boldsymbol{I}}
\global\long\def\Hb{\boldsymbol{H}}
\global\long\def\qb{\boldsymbol{q}}
 \global\long\def\hb{\boldsymbol{h}}
\global\long\def\tb{\boldsymbol{t}}
\global\long\def\rb{\boldsymbol{r}}
\global\long\def\wb{\boldsymbol{w}}
\global\long\def\mb{\boldsymbol{m}}
\global\long\def\eb{\boldsymbol{e}}
\global\long\def\qb{\boldsymbol{q}}
\global\long\def\ob{\boldsymbol{o}}
\global\long\def\ab{\boldsymbol{a}}
\global\long\def\argmax{\operatornamewithlimits{argmax}}

\global\long\def\argmin{\operatornamewithlimits{argmin}}

\section{Introduction}

\input{intro.tex}

\section{Related Work \label{sec:Related-Work}}

\input{related.tex}

\section{Methods\label{sec:Methods}}

\input{method.tex}

\section{Experimental Results \label{sec:Results}}

\input{exp_v2.tex}

\section{Discussion \label{sec:Discussion}}

\input{discuss.tex}
\bibliographystyle{plain}

\end{document}

%% file: abs.tex
Modern healthcare is ripe for disruption by AI. A game changer would
be automatic understanding the latent processes from electronic medical
records, which are being collected for billions of people worldwide.
However, these healthcare processes are complicated by the interaction
between at least three dynamic components: the illness which involves
multiple diseases, the care which involves multiple treatments, and
the recording practice which is biased and erroneous. Existing methods
are inadequate in capturing the dynamic structure of care. We propose
Resset, an end-to-end recurrent model that reads medical record
and predicts future risk. The model adopts the algebraic view in that
discrete medical objects are embedded into continuous vectors lying
in the same space. We formulate the problem as modeling sequences
of sets, a novel setting that have rarely, if not, been addressed.
Within Resset, the bag of diseases recorded at each clinic visit
is modeled as function of sets. The same hold for the bag of treatments.
The interaction between the disease bag and the treatment bag at a
visit is modeled in several, one of which as residual of diseases
minus the treatments. Finally, the health trajectory, which is a sequence
of visits, is modeled using a recurrent neural network. We report
results on over a hundred thousand hospital visits by patients suffered
from two costly chronic diseases -- diabetes and mental
health. Resset shows promises in multiple predictive tasks such as readmission prediction, treatments recommendation and diseases progression.

%% file: intro.tex
After stunning successes in cognitive domains, deep learning is expected
to transform healthcare \cite{tran2017deep}. Most remarkable results
thus far in health have been in diagnostic imaging \cite{esteva2017dermatologist,gulshan2016development},
which is a natural step given record\textendash breaking results in
computer vision. However, diagnostic imaging is only a small part
of the story. A full intelligent medical system should be able to
reason about the past (historical illness), present (diagnosis) and
future (prognosis). Here we adopt the notion of \emph{reasoning} as
``algebraically manipulating previously acquired knowledge in order
to answer a new question'' \cite{bottou2014machine}. For that we
learn to embed discrete medical objects into continuous vectors, which
lend themselves to a wide host of powerful algebraic and statistical
operators \cite{choi2016multi,tran2015learning}. For example, with
diseases represented as vectors, computing disease-disease correlation
is simply a cosine similarity between the two corresponding vectors.
Illness \textendash{} recorded as a bag of discrete diseases \textendash{}
can then be a function of set of vectors. The same holds for care.
Importantly, if diseases, treatments (or even doctors) are embedded
in the same space then recommendation of treatments (or doctors) for
a given disease will be as simple as finding the nearest vectors.

The algebraic view makes it easily to adapt powerful tools from the
recent deep learning advances \cite{lecun2015deep} for healthcare.
In particular, we can build end-to-end models for risk prediction
without manual feature engineering \cite{cheng2016risk,nguyen2016deepr,pham2017predicting,ravi2017deep}.
As the models are fully differentiable, credit assignment to distant
risk factors can be carried out \cite{nguyen2016deepr}, making the
models more transparent than commonly thought. The learning path through
recorded medical data necessitates the modeling of the dynamic interaction
between the three processes: the illness, the care and the data recording
\cite{pham2017predicting}. For the purpose of this paper, we assume
that a clinic visit at a time manifests through a set of discrete
diseases and treatments. A healthcare trajectory is therefore a sequence
of time-stamped records. This necessitates a set-theoretic treatment
of each visit and a dynamic treatment of the entire trajectory.

We introduce $\model$, a recurrent model of healthcare trajectory
as a \emph{sequence of sets}. Although neural modeling of unordered
sets has been recently studied \cite{vinyals2015order,zaheer2017deep},
sequences of sets have not been formally investigated, to the best
of our knowledge. In $\model$, features for a set are computed through
a multi-valued set function, which is permutation invariant. There
are two set functions, one for diseases and the other for treatments.
A dual-input function of these two set functions encodes the multi-disease\textendash multi-treatment
interaction at the visit level. Finally, visits are connected through
a LSTM to model the temporal dynamics between visits. 

With this design, $\model$ addresses an important aspect of healthcare:
the dynamic interaction between illness and care. Although care is
supposed to lessen the illness, it is often designed through highly
controlled trials where one treatment is targeted at one disease,
on a specific cohort, at a specific time \cite{kent2007limitations}.
Much less is known for the effect of multiple treatments on multiple
diseases, in general hospitalized patients, over time. A recent model
known as DeepCare \cite{pham2017predicting} partly addresses this
problem by considering the moderation effect of treatments on illness
state transition \emph{between visits}. However, unlike $\model$,
DeepCare does not address the multi-disease\textendash multi-treatment
interaction \emph{within visits}.

We evaluate $\model$ on the task of predicting the important medical
outcomes such as unplanned readmission or death at discharge, treatment
recommendation and future diseases. We focus on chronic diseases (diabetes
and mental health) as they are highly complex with multiple causes,
often associated with multiple comorbidities, and the treatments are
not always effective. Results from over one hundred thousand visits
to a large regional hospital data demonstrate the efficacy of $\model$.

To summarize, we claim the following contributions: 1) A novel representation
of time-stamped healthcare trajectory as a sequence of sets. 2) A
novel deep learning architecture for sequence of sets called $\model$,
which \textendash{} when applied to healthcare \textendash{} uncovers
the structure of the disease/treatment space and predicts future outcomes.
3) An evaluation of these claimed capabilities on real patients with
hundred thousands of hospital visits on three tasks: readmission prediction,
treatments recommendation and diseases progression.

The rest of the paper is organized as follows. Section~\ref{sec:Related-Work}
briefly review related work in healthcare and deep learning. Section~\ref{sec:Methods}
presents $\model$ in context of healthcare trajectory modeling. Experimental
results are reported in Section~\ref{sec:Results}. Finally, Section~\ref{sec:Discussion}
concludes the paper.

%% file: related.tex
\subsection{Deep Learning for Healthcare}

The past few years have witnessed an intense interest in applying
recent deep learning advances to healthcare \cite{ravi2017deep}.
The most ready area is perhaps medical imaging \cite{greenspan2016guest}.
Thanks to the record-breaking successes in convolutional nets in computer
vision, we now can achieve diagnosis accuracy comparable with experts
in certain sub-areas such as skin-cancer \cite{esteva2017dermatologist}.
However, it is largely open to see if deep learning succeeds in other
areas where data are less well-structured and of lower quality such
as electronic medical records (EMR) \cite{pham2017predicting}.

Within EMRs, three set of techniques have been employed. The first
is finding distributed representation of medical objects such as diseases,
treatments and visits \cite{choi2016multi,tran2015learning}. The
techniques are not strictly deep but they offer a compelling algebraic
view of healthcare. The second group of techniques involve 1D convolutional
nets, which are designed for detecting short translation invariant
motifs over time \cite{cheng2016risk,nguyen2016deepr}. The third
group, to which this paper belongs, employs recurrent neural nets
to capture the temporal structure of care \cite{choi2016retain,pham2017predicting}.
For more comprehensive review of this highly dynamic research area
in recent time, we refer to \cite{shickel2017deep}. 

Predictive healthcare begs the question of modeling treatment effects
\emph{over time}. This has traditionally been in the realm of randomized
controlled trials. Our work here is, on the other hand, based entirely
on observational administrative data stored in Electronic Medical
Records.

\subsection{Neural Nets for Sets}

Sets are fundamental to mathematics, and are pervasive in many learning
tasks such as clustering (set membership assignment), feature selection
(subset of features), multi-label learning (subset of labels) and
multi-instance learning (set classification). Due to its permutation-invariance
and variation in size, set does not lend itself naturally to traditional
neural networks. In \cite{vinyals2015order}, mapping set to set is
framed as mapping sequence to sequence, in which a set is ``pretended''
to be a pseudo-sequence. This does not address the permutation-invariant
property of sets. A more systematic investigation is \cite{zaheer2017deep},
where conditions for set functions are specified. Sets have also been
studied indirectly, e.g., in pooling operations in CNN; in attention
mechanism (e.g., see \cite{vaswani2017attention}) \textendash{} which
is essentially a function over sets, in deep multi-X learning \cite{pham2017one}.
Predicting sets have also been studied in \cite{hamid2017deepsetnet}.
In the existing literature, neural models of sequences of sets seem
to be missing.

%% file: method.tex
\begin{figure*}
\begin{centering}
\begin{tabular}{|c|}
\hline 
\includegraphics[width=1\textwidth]{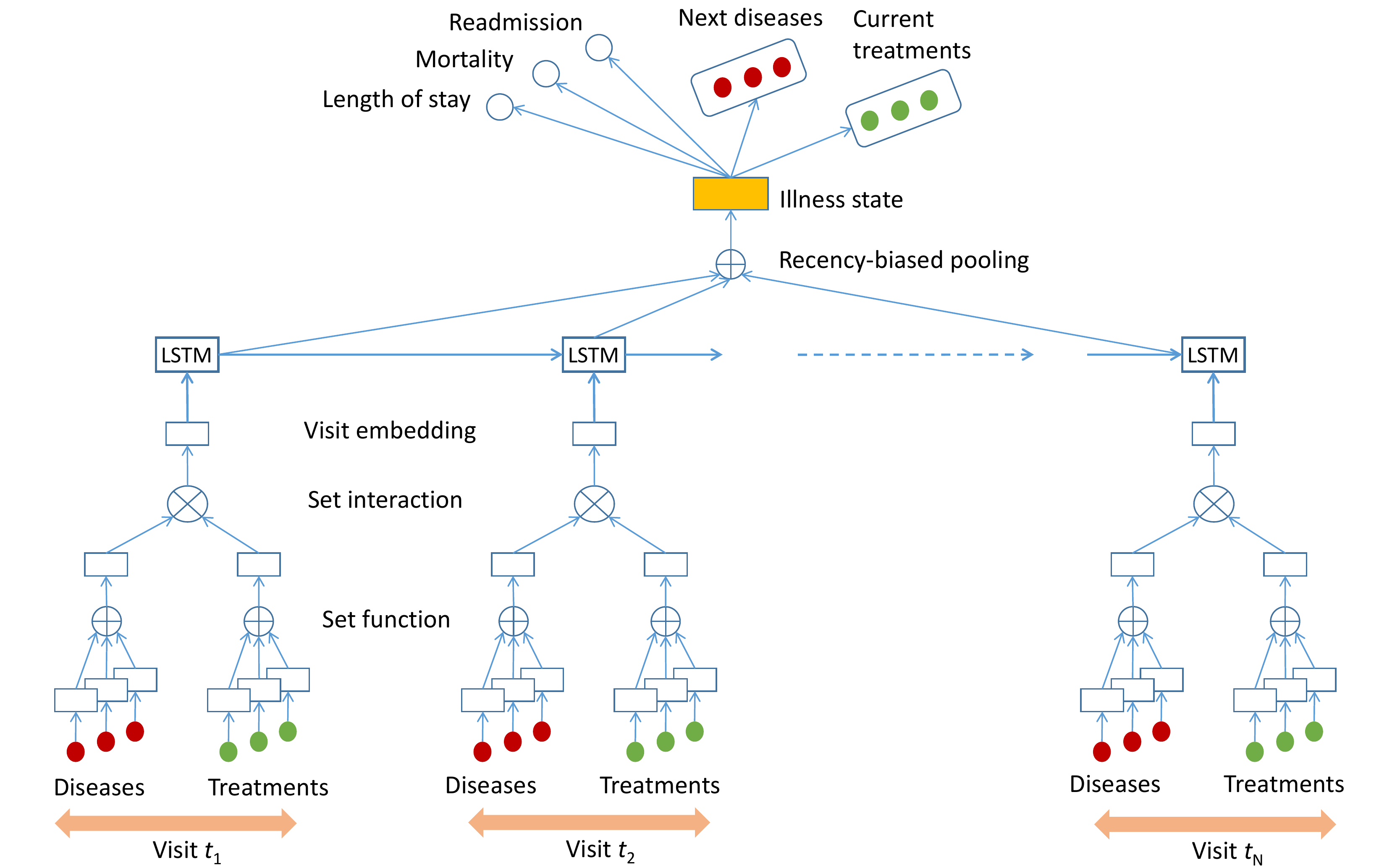}\tabularnewline
\hline 
\end{tabular}
\par\end{centering}
\caption{$\protect\model$ for sequence of sets, as a model of healthcare trajectory.\label{fig:model}}
\end{figure*}

In this section, we present our main modeling contribution, the $\model$
as a recurrent model for \emph{sequence of sets}, in the context of
its primary application in healthcare.

\subsection{Set Function}

Let us start with set, an unordered collection of elements. Let $S=\left\{ \eb_{1},\eb_{2},...\right\} $
be a set of vectors in $\mathbb{R}^{n}$. A set function $f(S)$ is
a mapping invariant against permutation of set elements, that is,
$f(S)=f(\pi S)$, for any permutation operator $\pi$. For simplicity,
we are interested in a function $f(S)$ that receives a set of vectors
$S$ and returns another vector in the same real space $\mathbb{R}^{n}$.
We use the following normalized set function:

\begin{align}
f(S) & \leftarrow\frac{\bar{\eb}_{S}}{\epsilon+\left\Vert \bar{\eb}_{S}\right\Vert }\quad\text{where}\quad\bar{\eb}_{S}=\max\left(\mathbf{0},\sum_{i\in S}\eb_{i}\right)\label{eq:set-func}
\end{align}
where $\epsilon>0$ is a smoothing factor. This is essentially a linear
rectifier of the sum, approximately normalized to unit vector. The
factor $\epsilon$ lets $\left\Vert f(S)\right\Vert \rightarrow0$
when$\left\Vert \bar{\eb}_{S}\right\Vert \rightarrow0$, but $\left\Vert f(S)\right\Vert \rightarrow1$
when $\left\Vert \bar{\eb}_{S}\right\Vert \gg0$.

\subsection{Clinic Visit as Set-Set Interaction \label{subsec:Clinic-Visit}}

Each medical record contains information about the history of clinic
visits by a patient. For simplicity, we consider a visit record as
a bag of diseases deemed relevant for care at the time of visit, and
a bag of treatments administered for the patient. Among older cohorts,
non-singleton bags are prevalent, reflecting the comorbidity picture
of modern healthcare, that is, an elderly typically suffers from multiple
co-occurring conditions. As a result, treatments must be carefully
administered to work with, or at least not to cancel out, each other.
This also calls for a sensible way to model the complexity of \emph{multi-disease\textendash multi-treatment
interaction}. Most existing bio-statistics methods, however, are designed
for simplified treatment effect against just one condition in a controlled
experimental setting, and thus inadequate in this operational setting.
Our solution is based on \emph{set-set interaction}, which we detail
subsequently.

We first use vector representation of diseases and treatments, following
the recent practice in NLP (e.g., see \cite{collobert2011natural}).
Let $\eb_{d}$ be the \emph{embedding} of disease $d$, $\eb_{p}$
the representation of the treatment $p$, and the vectors are embedded
in a common space. The bags of diseases and bags of treatments are
also represented as vectors in the same space. 

Denote by $D_{t}$ the bag of diseases and $I_{t}$ the bag of treatments
recorded for the visit at time $t$. Let $\db_{t}=f\left(D_{t}\right)$
be the set representation of the disease bag, and $\pb_{t}=f\left(I_{t}\right)$
the set representation of the treatment bag, as given in Eq.~(\ref{eq:set-func}).
Let $\vb_{t}=g\left(\db_{t},\pb_{t}\right)$ denote the interaction
function between diseases, as encapsulated in $\db_{t}$, and treatments,
as coded in $\pb_{t}$. A popular method is to use a bilinear function,
e.g., $\vb_{t}^{k}=\rho\left(\db_{t}^{\top}W^{k}\pb_{t}\right)$ where
$W^{k}$ is a matrix and $\rho$ is an element-wise nonlinear transformation,
but this will result in lots of parameters. 

One intuitive function is as follows:

\begin{equation}
g\left(\db_{t},\pb_{t}\right)=\rho\left(\Delta\right)\quad\text{where}\quad\Delta=\db_{t}-\pb_{t}\label{eq:disease-treatment-effect-subtractive}
\end{equation}
which we dub the \emph{subtractive} interaction. The difference $\Delta$
reflects the intuition that treatments are supposed to lessen the
illness. We found $\rho\left(\Delta\right)=\left(1+\Delta\right)^{2}$
works well, suggesting that the disease-treatment interaction is nonlinear.
This warrants a deeper investigation in future work, e.g., $\rho$
as a neural network itself. We also experimented with other interaction
forms: the \emph{implicit} with $\Delta=f\left(I_{t}\cup D_{t}\right)$,
the \emph{additive} with $\Delta=\db_{t}+\pb_{t}$, and \emph{multiplicative}
with $\Delta=\db_{t}\ast\pb_{t}$. See Section~\ref{sec:Results}
for empirical results.

\subsection{Healthcare Trajectory as Sequence of Sets}

While we might expect that the disease subset together with the treatment
subset reflect the illness state at the time of discharge, it is not
necessarily the case. This is because of several reasons. First, the
coding of those diseases and treatments is often optimized for billing
purposes, not all diseases are included. Second, errors do occur sometimes.
And third, the treatments usually take time to get the full intended
effect.

For this reason, it is better to include historical visits to assess
the current state as well as to predict future risk. An efficient
way is to model the visit sequences as a Recurrent Neural Network
(RNN) \cite{elman1990finding}. In this paper, we choose Long Short-Term
Memory (LSTM) due to its capability to remember distant events \cite{hochreiter1997long}.
Since each visit is a set \textendash{} or precisely, an interaction
of two sets \textendash{} \emph{health trajectory can be modeled as
a sequence of sets}. We term the model $\model$, which stands for
\textbf{Re}current \textbf{S}equence of\textbf{ Set}s. Figure~\ref{fig:model}
depicts a graphical illustration of $\model$.

Given a sequence of input vectors (one per visit) the LSTM reads an
input $\vb_{t}$ at a time and estimates the illness state $\hb_{t}$.
To connect to the past, LSTM maintains an internal short-term memory
$\cb_{t}$, which is updated after seeing the input. Let $\tilde{\cb}_{t}$
be the new candidate memory update after seeing $\vb_{t}$, the memory
is updated over time as:
\begin{align*}
\cb_{t} & =\fb_{t}\ast\cb_{t-1}+\ib_{t}\ast\tilde{\cb}_{t}
\end{align*}
where $\fb_{t}\in(\boldsymbol{0},\boldsymbol{1})$ is forget gate
determining how much of past memory to keep; $\ib_{t}\in(\boldsymbol{0},\boldsymbol{1})$
is the input gate controlling the amount of new information to add
into the present memory. The input gate is particularly useful when
some recorded information is irrelevant to the final prediction tasks.

The memory gives rise to the state as follows:

\[
\hb_{t}=\ob_{t}\ast\tanh(\cb_{t})
\]
where $\ob_{t}\in(\boldsymbol{0},\boldsymbol{1})$ is the output gate,
determining how much external information can be extracted from the
internal memory. The candidate memory and the three gates are parametric
functions of $(\vb_{t},\hb_{t-1})$.

With this long short-term memory system in place, information of the
far past is not entirely forgotten, and credit can be assigned to
it. Second, partially recorded information can be integrated to offer
a better picture of current illness.

\subsubsection{Regularizing state transitions}

For chronic diseases, it might be beneficial to regularize the state
transition. We consider adding the following regularizers to the loss
function:
\[
\frac{\beta}{T}\sum_{t=2}^{T}\left(\left\Vert \hb_{t}\right\Vert _{2}-\left\Vert \hb_{t-1}\right\Vert _{2}\right)^{2}
\]
as suggested in \cite{krueger2015regularizing}. This asks the amount
of information available at each time step, encapsulated in the norm
$\left\Vert \hb_{t}\right\Vert _{2}$, to be stable over time. This
is less aggressive than maintaining state coherence, i.e., by minimizing
$\frac{\beta}{T}\sum_{t=2}^{T}\left\Vert \hb_{t}-\hb_{t-1}\right\Vert _{2}^{2}$.

\subsection{Predictions}

Once the LSTM is specified, its states are pooled for prediction at
each admission or discharge, i.e., $\bar{\hb}_{t}=\text{pool}(\hb_{1:t})$.
The pooling function can be as simple as the \texttt{mean()}(i.e.,
$\bar{\hb}_{t}=\frac{1}{t}\sum_{j=1}^{t}\hb_{j}$) or\texttt{ last()}(i.e.,
$\bar{\hb}_{t}=\hb_{t}$). We also experimented with exponential smoothing,
i.e., 
\[
\bar{\hb}_{t}=\alpha\bar{\hb}_{t-1}+(1-\alpha)\hb_{t}
\]
for $\bar{\hb}_{1}=\hb_{1}$ and $\alpha\in[0,1]$. A small $\alpha$
would mean the recent visits have more influence in future outcomes.
Next, a differentiable classifier (e.g., a feedforward neural net)
is placed on top of the pooled state to classify the medical records
(e.g., those in population stratification) or to predict the outcome.
The loss function is typically the negative log-likelihood of outcome
given the historical observations, e.g., $-\sum_{t}\log P\left(y_{t}\mid\Db_{1:t},\Ib_{1:t}\right)$.
We emphasize here is the system is end-to-end differentiable, starting
from the disease and treatment lookup table at the bottom to the final
classifier at the top. No feature engineering is needed. 

There are many prediction tasks in healthcare. For example, at discharge
we can predict unplanned readmissions, mortality, future length-of-stay
within 12 months. These are single outcome prediction. In what follows,
we discuss two classes that predict multiple outcomes: Diseases prediction
and treatments recommendation.

\subsubsection{Diseases Prediction}

Disease prediction is an important task in healthcare. If the model
is presented with a sequence of admissions, it will learn to predict
what are the top $k$ probable diseases that the patient will have
in the next admission. This model shares similarity to the risk prediction
model at the recurrent level, however the top layers are reconstructed
to allow multiple label output. The pooled output $\bar{\hb}_{t}=\text{pool}(\hb_{1:t})$
from recurrent layer is used as feature input to the disease prediction
network. Finally, the output from this network are the top $k$ predicted
diseases: $\text{output}=k\text{argmax}\left(\text{net}(\bar{\boldsymbol{h}}_{t})\right)$.

Contrary to the single outcome model, we need to output multiple diseases
instead of binary value as in risk prediction. Hence we employed a
multi-label output approach to train the model. There are two ways
to define the loss function in this approach. The first one is to
let the network output the probabilities of every diseases by using
a sigmoid activation function. Then the loss function is simply binary
cross-entropy log loss. However, this method would suffer from the
imbalance between the small number of diseases a patient has and the
large number of all diseases, since it amplify the loss of the non-occurred
diseases by an amount proportional to the non-occurring ratio, hence
degrading the network performance. The second method is to only back-propagate
the loss of not picking up the right diseases in the next admission.
The network would output the probabilities of all diseases normalized
by a softmax function instead of per-disease probabilities by sigmoid
activation.

In this paper, we reported predicted diseases for $k=\left\{ 1,2,3\right\} $,
and used precision at $k$ as the performance measures.

\subsubsection{Treatments Recommendation}

Treatments recommendation is a very important task in healthcare.
It promises to reduce time and costs for doctors as well as patients,
and offers an unbiased access to care. This model is the same as disease
prediction model, but the top layer network is trained using the treatment
data from hospital instead of disease data.

%% file: exp_v2.tex
\subsection{Data}

We chose to study the data previously reported in \cite{pham2017predicting},
which consists of two chronic cohorts: diabetes and mental health.
These two cohorts are among the most prevalent, and have caused great
economical and societal burdens. As the natures of the two conditions
are very different (one is physical, the other is mental; one is typical
among the old, the other is typical among the young), consistent findings
will demonstrate the versatility of the models.

The data was collected between 2002-2013 from a large regional Australian
hospital. Each record contains at least 2 hospital visits. Diseases
and treatments are coded using the ICD-10 coding scheme. In ICD-10,
diseases are arranged in a tree, where the leaves represent the most
detailed sub-type classification. We use only the first two-level
in the ICD-10 tree to allow for sufficient statistics for each node.
Data statistics are summarized in Table~\ref{tab:Data-statistics}.
Overall, there are over a hundred thousand visits for both cohorts
combined. 

\begin{table}
\begin{centering}
\begin{tabular}{lcc}
\hline 
Statistics & Diabetes & Mental health\tabularnewline
\hline 
\# patients & 7,191 & 6,109\tabularnewline
\# visits & 53,208 & 52,049\tabularnewline
\% male & 55.5 & 49.4\tabularnewline
median age & 73 & 37\tabularnewline
\# diseases & 243 & 247\tabularnewline
\# treatments & 1,118 & 1,071\tabularnewline
\hline 
\end{tabular}
\par\end{centering}
\caption{Data statistics.\label{tab:Data-statistics}}
\end{table}

\subsection{Implementation}

Models are implemented in Julia using the Knet.jl package \cite{knet2016mlsys}.
Optimizer is Adam \cite{kingma2014adam} with learning rate of 0.01
and other default parameters. The recurrent layer hidden size is 32,
embedding size 32. The number of admissions is limited to the last
10 admissions. The RELU activation is applied at the input to the
recurrent layer. Dropout is used in training with dropout probability
0.5. The chosen exponential smoothing parameter after training is
0.1. Three baselines are implemented. One is bag-of-words trained
using regularized logistic regression (BoW+LR), where diseases and
treatments are considered as words, and the medical history as document.
No temporal information is modeled. Although this is a simplistic
treatment, prior research has indicated that BoW works surprisingly
well \cite{nguyen2016deepr,pham2017predicting}.

The other is a recent model known as Deepr \cite{nguyen2016deepr},
which is based on convolutional net for sequence classification. Unlike
the BoW, which are unordered, in Deepr words are sequenced by their
temporal order. Words of the same visit are randomly sequenced. Interaction
between diseases and treatments within a short period of time is partially
modelled through convolutional kernels. However, the Deepr does not
model the temporal transition between illness states but rather seeks
for the most risky states over the history. The Deepr model parameters
have embedding size 32, filter sizes 5, 10, 15 and the number of filters
is 60 (20 for each size). The Deepr input sequence length is limited
to the last 100 words. The last baseline is LSTM, which runs on the
same data as the Deepr.

\subsection{Predicting Unplanned Readmission}

Table~\ref{tab:AUC-all-methods} reports the Area Under the ROC Curve
(AUC) for all methods in predicting unplanned readmission. The proposed
methods shows a competitive performance against the baselines. The
$\model$ shows better prediction rate than those without set formulation
(the BoW, Deepr and LSTM). It suggests that a proper modeling of care
over time is needed, not only for understanding the underlying processes,
but also to achieve a competitive predictive performance.

\begin{table}
\begin{centering}
\begin{tabular}{lcc}
\hline 
\emph{Method} & \emph{Diabetes} & \emph{Mental health}\tabularnewline
\hline 
BoW+LR & 0.673 & 0.705\tabularnewline
Deepr \cite{nguyen2016deepr} & 0.680 & 0.714\tabularnewline
LSTM & 0.701 & 0.725\tabularnewline
\hline 
\textbf{$\model$} &  & \tabularnewline
\textbf{\textendash{} Implicit interaction} & \textbf{0.710} & \textbf{0.726}\tabularnewline
\textbf{\textendash{} Subtractive interaction} & \textbf{0.718} & \textbf{0.726}\tabularnewline
\textbf{\textendash{} Sub. interact + exp smoothing} & \textbf{0.701} & \textbf{0.730}\tabularnewline
\hline 
\end{tabular}
\par\end{centering}
\caption{Area Under the ROC Curve averaged over 5 folds in predicting unplanned
readmission. BoW = bag-of-words, LR = logistic regression. See Section~\ref{eq:disease-treatment-effect-subtractive}
for interaction modes of $\protect\model$.\label{tab:AUC-all-methods} }
\end{table}

\subsection{Treatment Recommendation}

Table~\ref{tab:Treatment-recommendation} reports the precision at
$k$ scores for different methods in predicting the treatments for
the diseases at the current time step. The top two scores in each
performance measure are shown in bold. In this task, no treatment
at the current admission is input to the model, only diseases input.
The table shows the proposed methods frequently have better performance
than the baselines. The additive and implicit interaction models show
better prediction rate than others for the diabetes cohort while the
subtractive and additive models outperform the remaining in mental
health data. The multiplicative model just performs similar to the
baseline on average. This suggests multiplicative interaction is a
too strong assumption. The exponential smoothing does not help improve
recommending treatments for mental health data.
\noindent \begin{flushleft}
\begin{table*}
\begin{centering}
\begin{tabular}{lcccccc}
\hline 
\multirow{2}{*}{\textbf{Method}} & \multicolumn{3}{c}{\textbf{Diabetes}} & \multicolumn{3}{c}{\textbf{Mental health}}\tabularnewline
\cline{2-7} 
 & P@1 & P@2 & P@3 & P@1 & P@2 & P@3\tabularnewline
\hline 
BOW+LR & 0.608 & 0.481 & 0.419 & 0.516 & 0.4382 & 0.395\tabularnewline
Deepr & 0.634  & 0.463 & 0.395 & 0.615 & 0.532 & 0.466\tabularnewline
LSTM & 0.694 & 0.535 & 0.446 & 0.614 & 0.507 & 0.427\tabularnewline
\hline 
$\model$ &  &  &  &  &  & \tabularnewline
\textendash{} Implicit interaction & \textbf{0.738} & \textbf{0.564} & \textbf{0.492} & 0.692 & 0.582 & \textbf{0.498}\tabularnewline
\textendash{} Additive interaction & \textbf{0.74} & \textbf{0.567} & \textbf{0.486} & \textbf{0.708} & \textbf{0.588} & 0.496\tabularnewline
\textendash{} Subtractive interaction & 0.704 & 0.553 & 0.48 & \textbf{0.7} & \textbf{0.591} & \textbf{0.51}\tabularnewline
\textendash{} Multiplicative interaction & 0.65 & 0.484 & 0.401 & 0.553 & 0.511 & 0.428\tabularnewline
\textendash{} Add. interaction with exp smoothing & 0.726 & \textbf{0.564} & 0.465 & 0.654 & 0.537 & 0.458\tabularnewline
\textendash{} Sub. interaction with exp smoothing & 0.730 & 0.561 & 0.465 & 0.641 & 0.528 & 0.452\tabularnewline
\hline 
\end{tabular}
\par\end{centering}
\caption{Treatment recommendation: Precision at $k$ averaged over 5 folds.
See Section~\ref{eq:disease-treatment-effect-subtractive} for interaction
modes of $\protect\model$. \label{tab:Treatment-recommendation}}
\end{table*}
\par\end{flushleft}

\subsection{Disease Prediction}
\noindent \begin{flushleft}
Table~\ref{tab:Disease-prediction} reports the precision at $k$
scores for predicting diseases in the next admission. Proposed methods
again frequently have better performance than the baselines. For this
task, the subtractive and implicit interaction models show better
prediction rate than others. The exponential smoothing clearly improves
the prediction rate for diabetes data. 
\par\end{flushleft}

\noindent \begin{flushleft}
\begin{table*}
\begin{centering}
\begin{tabular}{lcccccc}
\hline 
\multirow{2}{*}{\textbf{Method}} & \multicolumn{3}{c}{\textbf{Diabetes}} & \multicolumn{3}{c}{\textbf{Mental health}}\tabularnewline
\cline{2-7} 
 & P@1 & P@2 & P@3 & P@1 & P@2 & P@3\tabularnewline
\hline 
BOW+LR & 0.508  & 0.441 & 0.393 & 0.396 & 0.350 & 0.323\tabularnewline
Deepr & 0.496 & 0.42 & 0.397 & 0.424 & 0.392 & 0.346\tabularnewline
LSTM & 0.541 & 0.476 & 0.417 & 0.466 & 0.430 & 0.372\tabularnewline
\hline 
$\model$ &  &  &  &  &  & \tabularnewline
\textendash{} Implicit interaction & 0.530 & 0.478 & 0.438 & \textbf{0.504} & \textbf{0.471} & \textbf{0.406}\tabularnewline
\textendash{} Additive interaction & 0.528 & 0.496 & 0.449 & 0.488 & 0.448 & 0.392\tabularnewline
\textendash{} Subtractive interaction & 0.533 & 0.491 & 0.444 & \textbf{0.494} & \textbf{0.469} & \textbf{0.41}\tabularnewline
\textendash{} Multiplicative interaction & 0.496 & 0.44 & 0.401 & 0.453 & 0.406 & 0.362\tabularnewline
\textendash{} Add. interaction with exponential smoothing & \textbf{0.563} & \textbf{0.513} & \textbf{0.459} & 0.468 & 0.429 & 0.373\tabularnewline
\textendash{} Sub. interaction with exponential smoothing & \textbf{0.567} & \textbf{0.516} & \textbf{0.46} & 0.47 & 0.43 & 0.376\tabularnewline
\hline 
\end{tabular}
\par\end{centering}
\caption{Disease prediction: Precision at $k$ averaged over 5 folds. See Section~\ref{eq:disease-treatment-effect-subtractive}
for interaction modes of $\protect\model$.\label{tab:Disease-prediction}}
\end{table*}
\par\end{flushleft}

\subsection{Visualization}

\begin{figure*}[!t]
\centering{}%
\begin{tabular}{cc}
\includegraphics[width=0.48\textwidth,height=0.5\columnwidth]{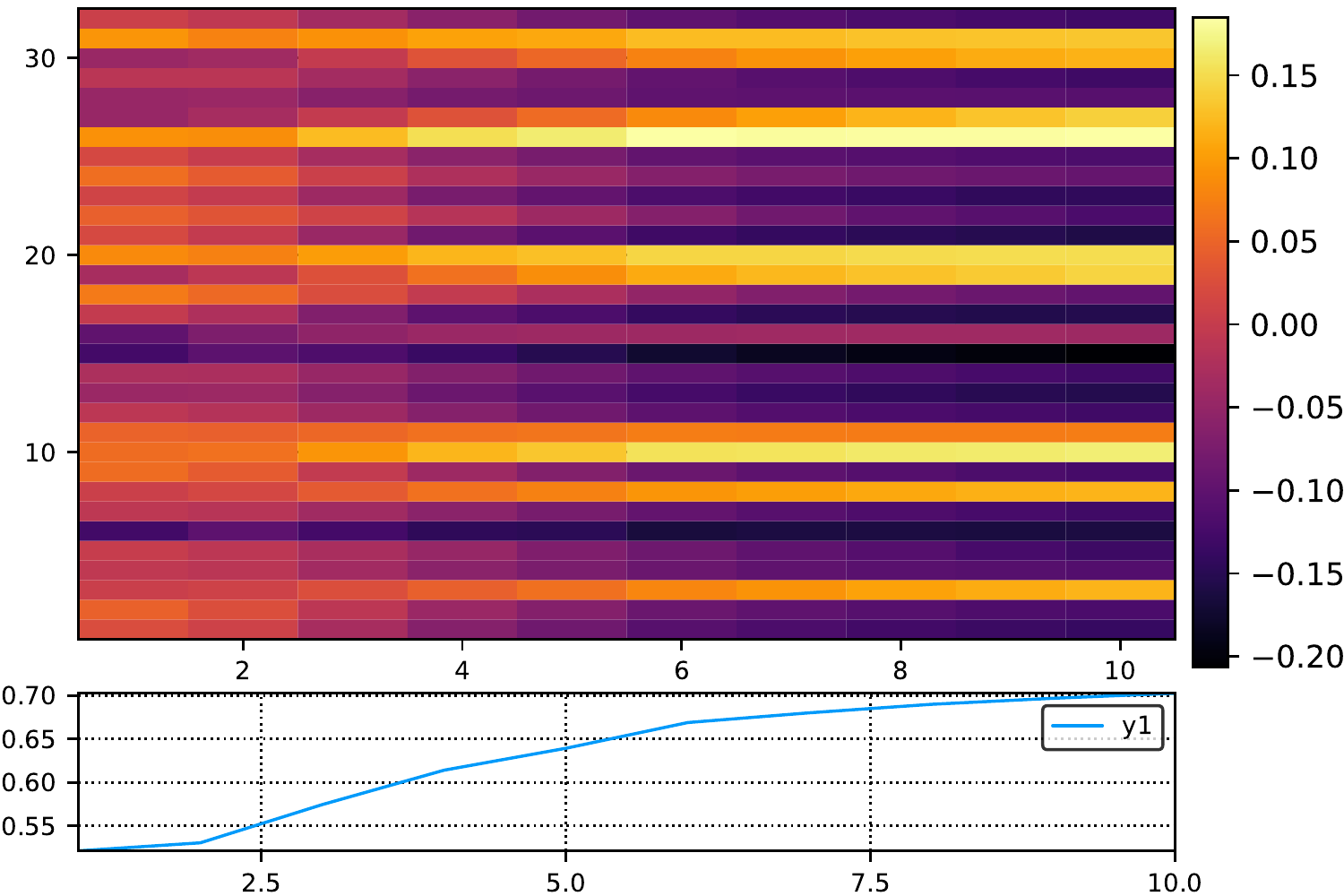} & \includegraphics[width=0.48\textwidth,height=0.5\columnwidth]{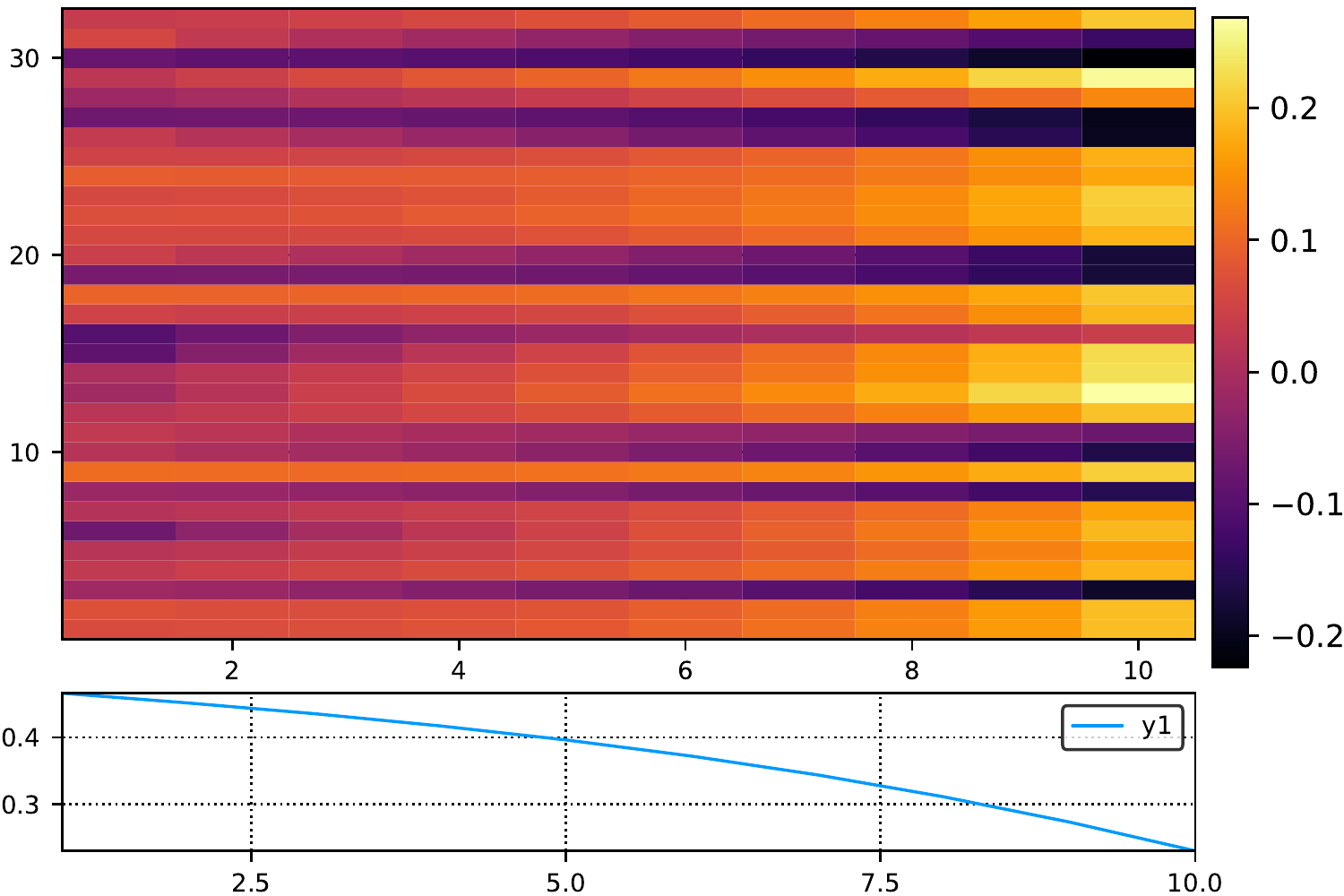}\tabularnewline
(a) Worsening progression ($P=0.70$) & (b) Improving progression ($P=0.23$)\tabularnewline
\end{tabular}\caption{Illness state progression over time, measured as $\protect\hb_{t}$
for the last 10 visits. \textbf{Left figure:} a high-risk case with
70\% chance of readmission at time T=10. \textbf{Right figure}: a
low-risk case with 23\% chance at the end of the sequence. Best viewed
in color.\label{fig:hidden-states}}
\end{figure*}

Visualization is of paramount importance in healthcare because of
the demand for transparency. The progression of the illness state
and probability of readmission over time is visualized in Fig.~\ref{fig:hidden-states}
for two typical patients. The high-risk case is shown in Fig.~\ref{fig:hidden-states}(a)
\textendash{} it seems that the illness gets worse over time. In contrast,
the low-risk case is depicted in Fig.~\ref{fig:hidden-states}(b),
where the illness is rather stable over time.

\begin{figure*}[!t]
\centering{}%
\begin{tabular}{cc}
\includegraphics[width=0.48\textwidth,height=0.5\columnwidth]{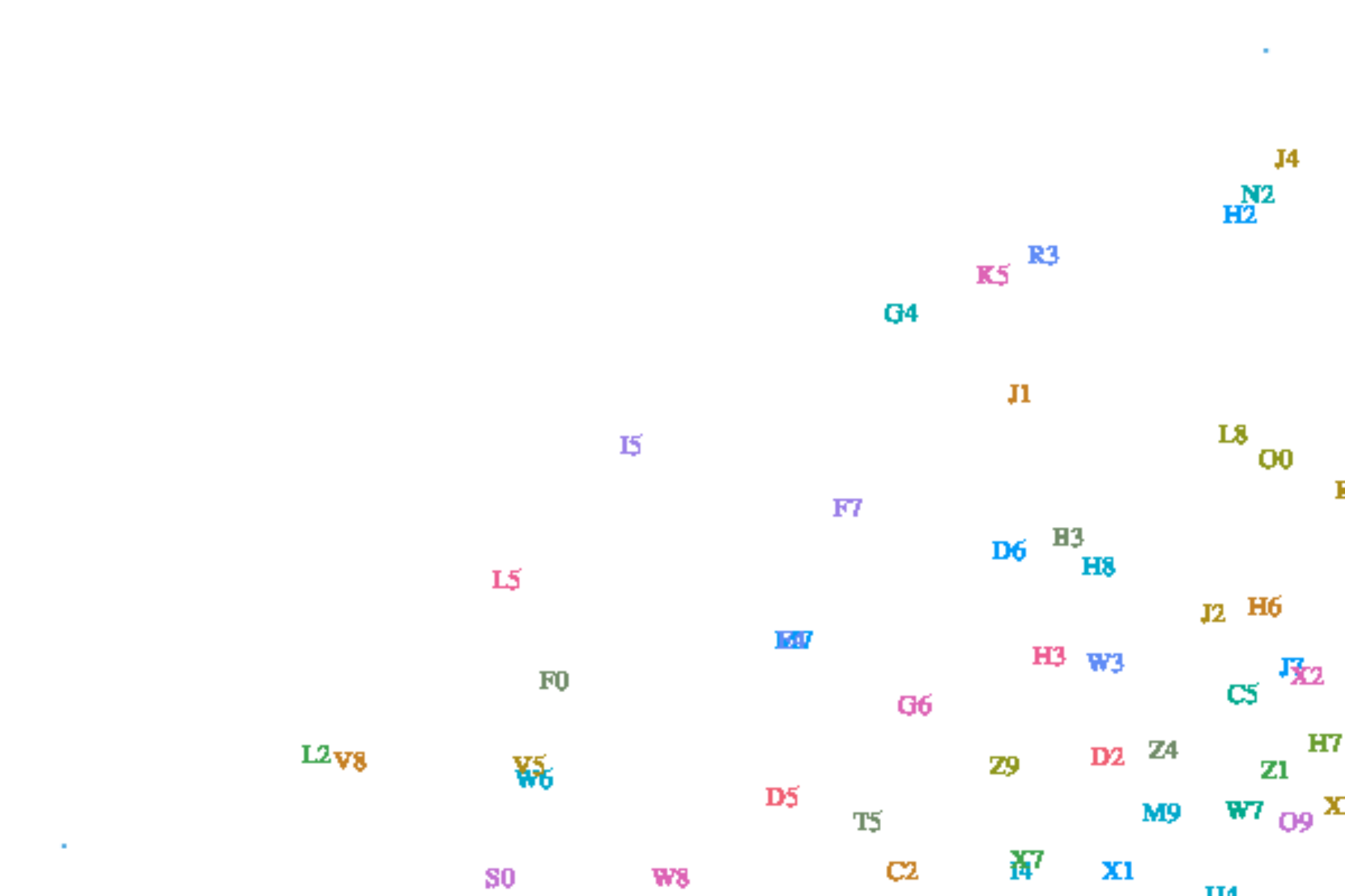} & \includegraphics[width=0.48\textwidth,height=0.5\columnwidth]{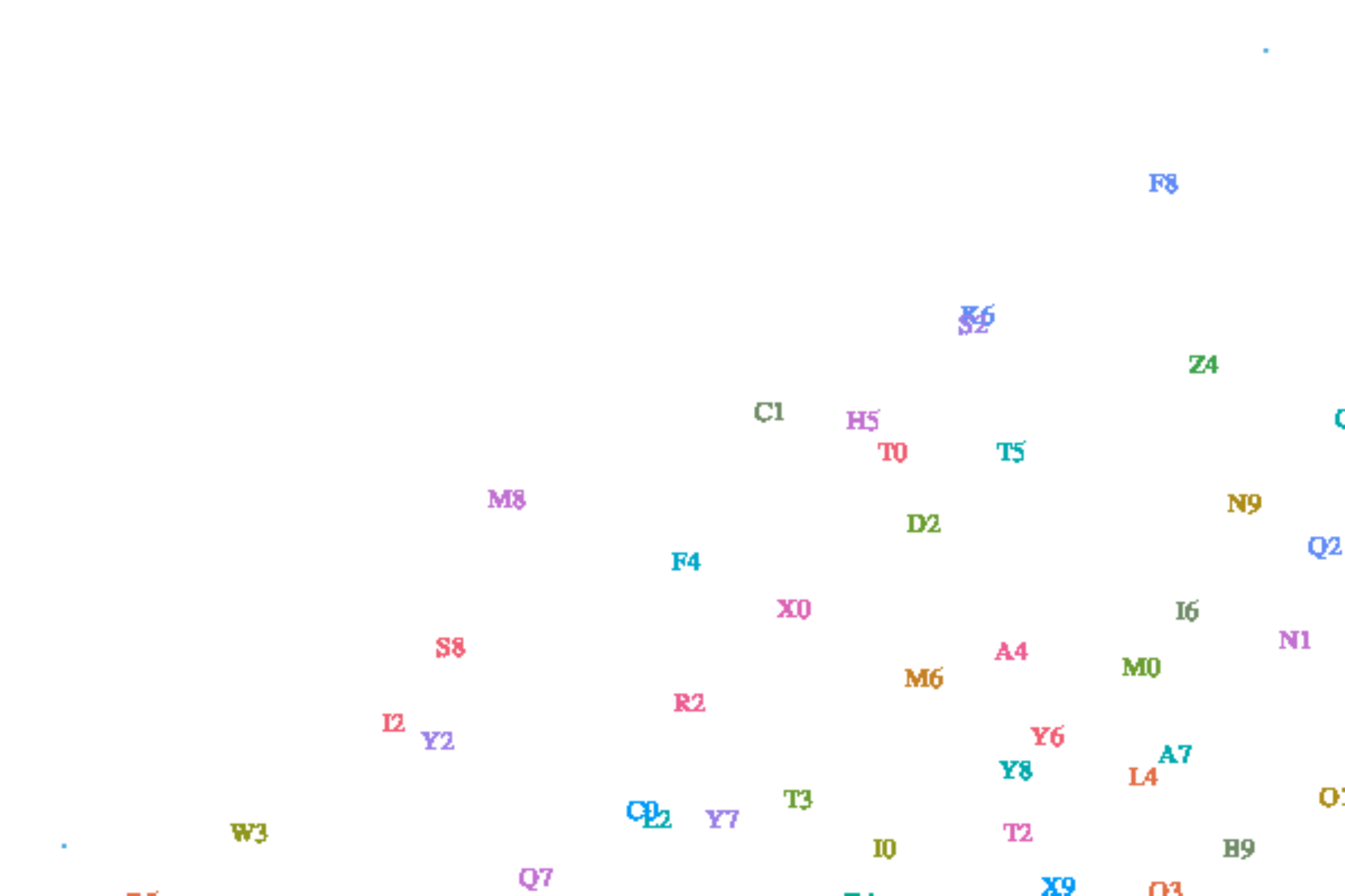}\tabularnewline
(a) Diabetes related diseases & (b) Mental health related\tabularnewline
\end{tabular}\caption{Projection of ICD-10 diagnostic codes using t-SNE after embedding.
Best viewed in color. Interested readers are referred to the WHO official
ICD-10 scheme book {[}http://apps.who.int/classifications/icd10/browse/2010/en{]}.\label{fig:disease-embedding}}
\end{figure*}

Code embedding also reveals the space of diseases, as visualized in
Fig.~\ref{fig:disease-embedding}.

%% file: discuss.tex
We have taken an \emph{algebraic view of healthcare} in that medical
artifacts are represented as algebraic objects such as vectors and
tensors. The continuous representation of diseases make it easy to
study the disease space, that is, which diseases are related and may
be interacting. The same holds for the treatments, and the clinic
visits. The view also allows natural modelling of the evolution of
illness as a result of the interaction between multiple diseases and
multiple treatments over time. More specifically, we have argued for
representing a healthcare trajectory as a sequence of (interaction
of) sets, which is then realized by our new model dubbed $\model$.
The model employs a simple multi-valued set function for diseases
and for treaments. Multi-disease\textendash multi-treatment interaction
per visit is a dual-input function of the two set functions. A healthcare
trajectory is then modelled using LSTM for its capability of memorizing
distant events. Importantly, the entire system is \emph{end-to-end}:
the model reads the medical record and predicts future risks \emph{without}
any manual feature engineering. Results on over a hundred thousand
visits by patients suffering from chronic conditions, diabetes and
metal health, demonstrate the usefulness of the model.

Future work will refine $\model$ in the healthcare context to address
the irregular timing of visits, more comprehensive set functions (e.g.,
with self-attention), interaction functions, and predicting sequence
of sets. We wish to emphasize here that $\model$ can be tailored
to other problems of sequence of sets. For example, a video is a sequence
of shots, each of which is a set of objects and actions. 